%% file: main.tex
\documentclass[letterpaper, 10 pt, conference]{ieeeconf}  % Comment this line out if you need a4paper

\IEEEoverridecommandlockouts                              % This command is only needed if 
\usepackage{float}
\usepackage{graphics} % for pdf, bitmapped graphics files
\usepackage{epsfig} % for postscript graphics files
\usepackage{mathptmx} % assumes new font selection scheme installed
\usepackage{times} % assumes new font selection scheme installed
\usepackage{amsmath} % assumes amsmath package installed
\usepackage{amssymb}  % assumes amsmath package installed
\usepackage[subpreambles=true]{standalone}
\usepackage{import}
\usepackage{wrapfig,lipsum}
\usepackage{graphicx}
\usepackage{booktabs}
\usepackage{multirow}
\usepackage{placeins}
\usepackage{gensymb}
\usepackage{xcolor}
\usepackage{soul}
\usepackage{url}
\setstcolor{red}
\usepackage[normalem]{ulem}               % to striketrhourhg text
\newcommand\redout{\bgroup\markoverwith
{\textcolor{red}{\rule[0.5ex]{2pt}{0.8pt}}}\ULon}

\restylefloat{table}
\usepackage[ruled,linesnumbered, noend]{algorithm2e}

\SetCommentSty{mycommfont}

\newenvironment{myitem}{\begin{list}{$\bullet$}
{\setlength{\itemsep}{-0pt}
\setlength{\topsep}{0pt}
\setlength{\labelwidth}{0pt}
\setlength{\leftmargin}{10pt}
\setlength{\parsep}{-0pt}
\setlength{\itemsep}{0pt}
\setlength{\partopsep}{0pt}}}%
{\end{list}}

\title{\Large \bf
Robust, Occlusion-aware Pose Estimation for Objects Grasped by Adaptive Hands
}

\author{Bowen Wen, Chaitanya Mitash, Sruthi Soorian, Andrew Kimmel, Avishai Sintov and Kostas E. Bekris%<-this stops a space
\thanks{The authors are with the Computer Science Dept. of
  Rutgers Univ. in NJ, USA.  This work is supported by NSF awards IIS-1734492, IIS-1723869, CCF-1934924. The opinions \& findings in this paper do not necessarily reflect the sponsor's views. Email:
  \{bw344,kostas.bekris\}@cs.rutgers.edu.}%
}

\begin{document}

\maketitle
\thispagestyle{empty}
\pagestyle{empty}

%%%%%%%%%%%%%%%%%%%%%%%%%%%%%%%%%%%%%%%%%%%%%%%%%%%%%%%%%%%%%%%%%%%%%%%%%%%%%%%%
\begin{abstract}

\import{sections/}{abstract}

\end{abstract}

%%%%%%%%%%%%%%%%%%%%%%%%%%%%%%%%%%%%%%%%%%%%%%%%%%%%%%%%%%%%%%%%%%%%%%%%%%%%%%%%
\section{INTRODUCTION}

\import{sections/}{introduction}

%%%%%%%%%%%%%%%%%%%%%%%%%%%%%%%%%%%%%%%%%%%%%%%%%%%%%%%%%%%%%%%%%%%%%%%%%%%%%%%%
\section{RELATED WORK}
\import{sections/}{related_work}

%%%%%%%%%%%%%%%%%%%%%%%%%%%%%%%%%%%%%%%%%%%%%%%%%%%%%%%%%%%%%%%%%%%%%%%%%%%%%%%%
\section{PROBLEM FORMULATION}
\import{sections/}{problem_formulation}

%%%%%%%%%%%%%%%%%%%%%%%%%%%%%%%%%%%%%%%%%%%%%%%%%%%%%%%%%%%%%%%%%%%%%%%%%%%%%%%%
\section{APPROACH}
\import{sections/}{approach}

%%%%%%%%%%%%%%%%%%%%%%%%%%%%%%%%%%%%%%%%%%%%%%%%%%%%%%%%%%%%%%%%%%%%%%%%%%%%%%%%
\section{EXPERIMENTS}
\import{sections/}{experiments}

\section{CONCLUSIONS}

\import{sections/}{conclusion}


%%%%%%%%%%%%%%%%%%%%%%%%%%%%%%%%%%%%%%%%%%%%%%%%%%%%%%%%%%%%%%%%%%%%%%%%%%%%%%%%

\bibliographystyle{IEEEtran}
\bibliography{ref.bib}

\end{document}

%% file: sections/abstract.tex
Many manipulation tasks, such as placement or within-hand manipulation, require the object's pose relative to a robot hand. The task is difficult when the hand significantly occludes the object. It is especially hard for adaptive hands, for which it is not easy to detect the finger's configuration. In addition, RGB-only approaches face issues with texture-less objects or when the hand and the object look similar. This paper presents a depth-based framework, which aims for robust pose estimation and short response times. The approach detects the adaptive hand's state via efficient parallel search given the highest overlap between the hand's model and the point cloud. The hand's point cloud is pruned and robust global registration is performed to generate object pose hypotheses, which are clustered. False hypotheses are pruned via physical reasoning. The remaining poses' quality is evaluated given agreement with observed data. Extensive evaluation on synthetic and real data demonstrates the accuracy and computational efficiency of the framework when applied on challenging, highly-occluded scenarios for different object types. An ablation study identifies how the framework's components help in performance. This work also provides a dataset for in-hand 6D object pose estimation. Code and dataset are available at: \url{https://github.com/wenbowen123/icra20-hand-object-pose}

%% file: sections/introduction.tex
\begin{figure}
  \centering
  \includegraphics[width=0.48\textwidth]{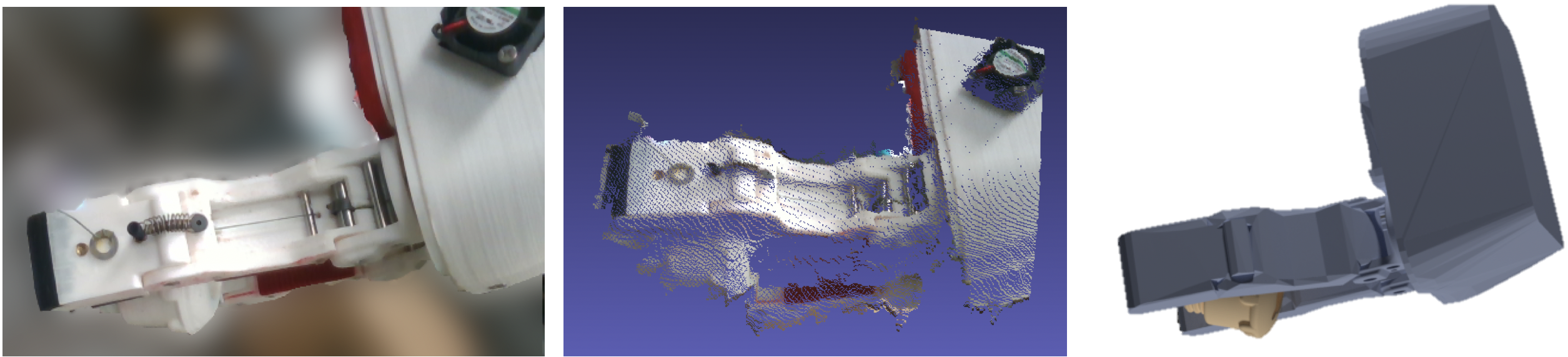}
  \vspace{-.3in}
  \caption{Left: Original image showing the adaptive hand  grasping and severely occluding a texture-less object. Middle: Point-cloud data.  Right: Scene reconstruction given the output of the approach. }
  \vspace{-.3in}
  \label{fig:intro}
\end{figure}

Robot manipulation often requires recognizing objects and detecting their 6D pose, i.e., position and orientation. Applications include logistics \cite{Correll:2016aa}, where picking is a frequent task. Once picked, an object may need to be purposefully placed for packaging, sorting or restocking. Depending on the task, regrasping or within-hand manipulation may also be required. These objectives need the object's 6D pose relative to the robot's hand post-grasp. Most existing work in pose estimation is focusing on the pre-grasp case \cite{zeng2017multi, schwarz2018rgb, liu2012fast, zhu2014single, mitash2019physics}, which is not always a good indicator of the post-grasp one due to the effects of  contact. This is especially true for adaptive hands, such as underactuated, compliant systems that naturally and safely adapt to an object's shape as in Fig. \ref{fig:intro}. There are multiple challenges that arise in this context: 

%Bottom Left: 3D models of the hand and object.

\noindent\textbf{-} \textbf{Severe occlusions:} The hand often significantly occludes the grasped object. Thus, solutions need to robustly distinguish the target object from the robot's fingers and noisy scene. Small objects further
complicate the process as they are mostly covered by the hand from the camera's viewpoint.

\noindent\textbf{-} \textbf{Unpredictable contacts and dynamic tasks:} Pre-grasp pose estimation does not suffer as much from occlusions. Recent work for in-hand pose estimation \cite{choi2016using} assumes the pose does not change significantly upon grasping and can initialize ICP (Iterative Closest Point). But as the hand grasps the object, the pose changes dynamically. This is also true if regrasping or within-hand manipulation is performed, where it is difficult to account for contacts, especially for compliant and adaptive hands. 6D pose tracking \cite{issac2016depth, trinh2018modular, choi2013rgb, fallon2012efficient} can help but also requires a good initial estimate. If the tracking loses the object, robust pose estimation given a highly-occluded snapshot is still needed.

\noindent\textbf{-} \textbf{Robustness and Generalizability} Pose estimation based on color or texture data \cite{choi2016using, andrychowicz2018learning, xiang2017posecnn, tremblay2018deep, kokic2019learning} can be sensitive to lighting conditions, and challenging for texture-less objects or when the object and the robot hand look similar. Extracting local 3D descriptors and finding correspondences \cite{drost2010model,buch2017rotational,zeng20173dmatch} may suffer from limited object visibility.

\begin{figure*}[t]
  \centering
  \includegraphics[width=0.95\textwidth]{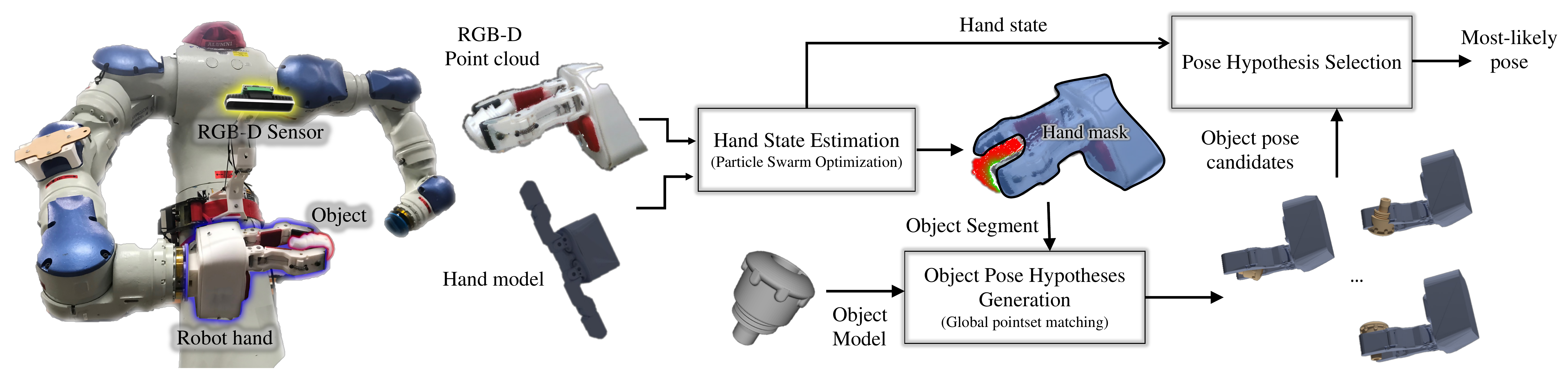}
  \vspace{-.2in}
  \caption{The framework acquires the RGB-D point cloud and computes the configuration of the adaptive hand given its CAD model. From this estimate, the hand is removed from the point cloud and the object segment is recovered. A set of pose candidates is generated by matching the segment to the object's model. The most likely pose is returned by evaluating the consistency of the interactions between the estimated hand and the in-hand object.   \vspace{-.1in}}
  \label{fig:pipeline}
\end{figure*}

%%% WE NEED TO MOTIVATE AND DESCRIBE ADAPTIVE HANDS

This paper presents a framework for robust, within-hand 6D object pose estimation using a consumer-level depth sensor. It addresses the issues arising from adaptive hands and focuses on the Yale Hand T42 \cite{odhner2013open}, given its use for dexterous manipulation \cite{Sintov2019LearningAS, bliefspacekimmel}.  A key feature is the estimation of the hand's state to help infer the object's region on the image. The method builds a hand-SDF (Signed Distance Field) to regularize the object's pose given physical constraints. This makes the task computationally manageable even under severe occlusions. The proposed framework exhibits these properties: 
\begin{myitem}
\item High precision; it achieves high accuracy, even for a tight error threshold of 5mm under the {\tt ADI} metric \cite{hinterstoisser2012model}.
\item Computational efficiency; as it returns the pose of the object and the state of the adaptive hand in 0.5 to 0.7 seconds. The hand's state estimate may also be helpful for control purposes;
\item Robustness; as the method works for various objects, including textureless ones, and with a cluttered background, where RGB-based methods would struggle.
\end{myitem}

%In particular, the framework efficiently searches in parallel the configuration space of the hand's fingers, given motion constraints. Finger configurations are found by matching the point cloud and the hand model. Given the hand's state, the point cloud is pruned and the remaining subset is assigned to the object. An accelerated version of global registration is performed, given the object's model, to generate a pool of pose hypotheses \cite{mitash2018robust, mellado2014super}. The hypotheses are clustered to identify a smaller set of pose candidates, which are further pruned via collision checking and depth-rendering comparison, given the hand state. Finally, the best pose candidate is computed given agreement with the point cloud.

%%We deliberately do not harness any grasping quality metric %%\cite{Miller2004GraspitAV} since the objective here is for any-time in-hand %%object state estimate that can take place during the whole manipulation process %%instead of a stable grasp.

This work also contributes a synthetic and a real dataset, where an adaptive hand holds various objects, with RGB-D data and ground truth information, since no related dataset exist in the literature beyond for objects contained in human hands \cite{issac2016depth}. Experiments on both datasets demonstrate the effectiveness, robustness and  efficiency of the proposed system for multiple objects in various scenarios, compared against state-of-the-art methods.  An ablation study highlights how the method's critical components help in performance.

%%the whole framework can be easily extended to other adaptive hands or general rigid grippers without loss of generality. 

%To our best knowledge, this is the first work that aims to solve the problem of %any-time single-frame 6D object pose estimation during in-hand manipulation for %adaptive hands. Both synthetic and real dataset used in this work will be released %so as to benefit the community.

%% file: sections/related_work.tex
This section covers different approaches for object pose estimation related to manipulation tasks.

\textbf{Alternatives to Vision:} Various sensors have been used for in-hand pose estimation, such as proprioception \cite{mason2012autonomous,homberg2015haptic},  and contact/force sensing \cite{tian2019manipulation,bimbo2016hand,aquilina2018principal}. Such sensors have also been combined with vision to decrease uncertainty \cite{schmidt2015depth,allen1988integrating,hebert2011fusion,zhang2012application,yu2018realtime,pfanne2018fusing,bimbo2013combining,chalon2013online}. Nevertheless, these sensing modalities are not always accessible, as they require careful engineering of the hands and increase cost. Under-actuated adaptive hands, for instance, do not often provide information for identifying finger configurations. Thus, a vision-only solution is desirable.

%\redout{visual exteroception \cite{issac2016depth,collet2011moped,bohg2017interactive,schmidt2015depth},}

\textbf{Single Image Object Pose Estimation:}
Recent advances in
object detection \cite{ren2015faster, redmon2016you} and pose
estimation \cite{xiang2017posecnn,mitash2019scene} have shown promise given access to sufficient labeled data. This allows to project an object's 3D bounding box on the image and solve for a pose using PnP \cite{tekin2018real, tremblay2018deep}. This is problematic, however, under severe occlusions. Alternatively, direct 6D pose regression has been attempted \cite{xiang2017posecnn, wang2019densefusion}. Nevertheless, the complexity of SO(3) results in instability in training and prediction. Recent work \cite{Kokic2019LearningTE,hasson2019learning} attempts to jointly estimate a human hand and the in-hand 6D object pose accounting for physical consistency but the resulting precision is not sufficient for manipulation. In contrast, a robotic hand's kinematic information is available, which helps increase precision.

%, inspired by 6D camera localization in SLAM \cite{kendall2015posenet}

\textbf{3D Registration Methods:} Registration \cite{drost2010model} often uses local geometry features followed by voting, which makes them sensitive to point cloud density that is problematic under severe occlusions. Alignment solutions can use gradient descent optimization \cite{zhou2016fast} but again degrade under severe occlusions, when only few features and correspondences can be extracted on the small point cloud segment of the object. Super4PCS has been shown effective in global registration, whereas its RANSAC nature makes it inefficient when large number of outliers exist. This work builds upon prior efforts \cite{zhou2016fast} and achieves higher accuracy with faster speed by introducing heuristics-guided sampling.
%It samples various 4-points co-planar bases on the source cloud and searches for the congruent sets on the target cloud in a \textit{RANSAC} manner. It generates a pool of rigid alignment hypotheses along with their scores measured by {\tt LCP}.

%In addition, the above approaches do not utilize physical constraints, which can %significantly help in robotic in-hand manipulation setups. 

\textbf{Object Pose Tracking:}  Methods have used a variety of approaches: GPU-accelerated particle filtering with a likelihood estimation based on color, distance and normals \cite{choi2013rgb}; modeling occlusions to eliminate outliers \cite{Wthrich2013ProbabilisticOT}; Gaussian Filtering to track objects using depth \cite{issac2016depth}. Promising precision is achieved for small errors but tracking loss arises frequently. Recent work \cite{deng2019poserbpf} formulates the 6D object pose tracking problem in the Rao-Blackwellized particle filtering framework. This method, however, requires a reliable single image pose for re-initialization upon tracking loss. The current work differs from the above in that it achieves fast, high precision estimates from individual high-occlusion snapshots without knowledge about previous frames. It can be integrated with such tracking frameworks to (re-)initialize.

\textbf{Visual Servoing:} A simple solution is to attach fiducial markers \cite{munoz2012aruco,yu2018realtime} on the object \cite{alli2018LearningMO, Cruciani2018DexterousMG, Sintov2019LearningAS} but it is not always practical to keep the marker visible, especially during in-hand manipulation. Additionally, complex surfaces make the attachment troublesome. Recent work trained an end-to-end policy network to perform within-hand manipulation while reasoning about object pose \cite{andrychowicz2018learning}. Computational resources, however, prevent it from easy application across conditions, such as objects unseen during training or having less distinctive features. Another effort estimated object pose by first segmenting the robot hand given a Naive Bayes classifier and then performing ICP (Iterative Closest Point) for the object segment, assuming the object does not move much upon grasping \cite{choi2016using}. This assumption is often violated when grasping or in-hand manipulation leads to object slippage. The current work does not depend on a pre-grasp estimate.

%%This work shows by jointly estimate the finger states and generate pose %%hypothesis of manipulating object based on fast global registration methods, %%physical consistency and scene-level reasoning together help to effectively %%reject ambiguous incorrect pose hypothesis, making high precision 6D object pose %%estimation possible in various challenging scenarios.

%% file: sections/problem_formulation.tex
% Given the object mesh $M$ and the depth image $I$ captured by a common RGB-D camera (e.g. Kinetic, RealSense), our goal is to estimate 6D pose of the in-hand object, i.e. the rigid transformation $T_M^{C}$ that brings the object mesh $M$ to the camera frame $C$. The whole system is shown in Fig. \ref{}. Camera is presumed to be calibrated and rough transformation from hand wrist frame $H$ to camera $T_H^{C}$ is obtained based on forward kinematics. In this work, we conduct all experiments using Yale Hand T42 \cite{odhner2013open} as a representative adaptive hand given its previous success in dexterous manipulation and publicly accessibility, while our system can also be extended to other kinds of adaptive hand or general rigid grippers without loss of generality. 

Given a depth image from camera $C$, a mesh model $M$ of object $O$, the goal is to compute $O$'s 6D pose, i.e., the rigid transform $T_M^{C}$, where $O$ is grasped by an adaptive hand in $C$'s view. The work considers under-actuated hands (the Yale Hand T42 \cite{odhner2013open}) for which a CAD model is available. The hand state determined by configuration of the $N$ fingers $x_H = \{q_{F_i}\}_{i=1}^N$ are initially unknown and not available. The camera is calibrated and the transform $T_H^{C}$ of the hand's wrist frame $H$ to the camera is available.

%% file: sections/approach.tex
% The proposed method approaches the problem by (1) parallelized evolutionary optimization for finger pose estimation on the adaptive hand, and (2) heuristics guided global registration running on non-hand region point cloud which generates a pool of 6D object pose hypothesis, followed by (3) scene-level physics reasoning accounting for interaction between estimated adaptive hand configuration and object for hypothesis rejection and selection.

Fig.~\ref{fig:pipeline} outlines the proposed approach with 3 key components: 1) parallel evolutionary optimization to estimate the hand's configuration; 2) heuristics-guided global pointset registration to generate pose hypotheses for the object; 3) scene-level physics reasoning that considers the hand-object interaction to find the most-likely object pose.

\subsection{Hand State Estimation} \label{sec_hand_pose}

% Joint hand and object pose estimation have shown success in multiple tasks related to human manipulation using deep neural networks \cite{Kokic2019LearningTE,hasson2019learning}. Similarly, 6D pose estimation of in-hand object during robot manipulation also benefits from explicit estimation of hand states by two folds: (1) it can help to localize object by inferring hand and object region on the image using estimated hand configuration \cite{choi2016using}; (2) SDF (Signed Distance Field) can be built from the estimated hand pose to strongly regularize the object pose estimation space by physical reasoning, making this challenging task approachable even under severe occlusions. Moreover, the accessibility to robot hand mesh along with the fingers' motion constraints enable searching methods to be carried out efficiently in the reduced state space. Therefore, full hand pose estimation is performed before estimating the 6D object pose.

\begin{wrapfigure}{r}{0.2\textwidth}
  \vspace{-.4in}
  \centering
  \includegraphics[width=0.15\textwidth]{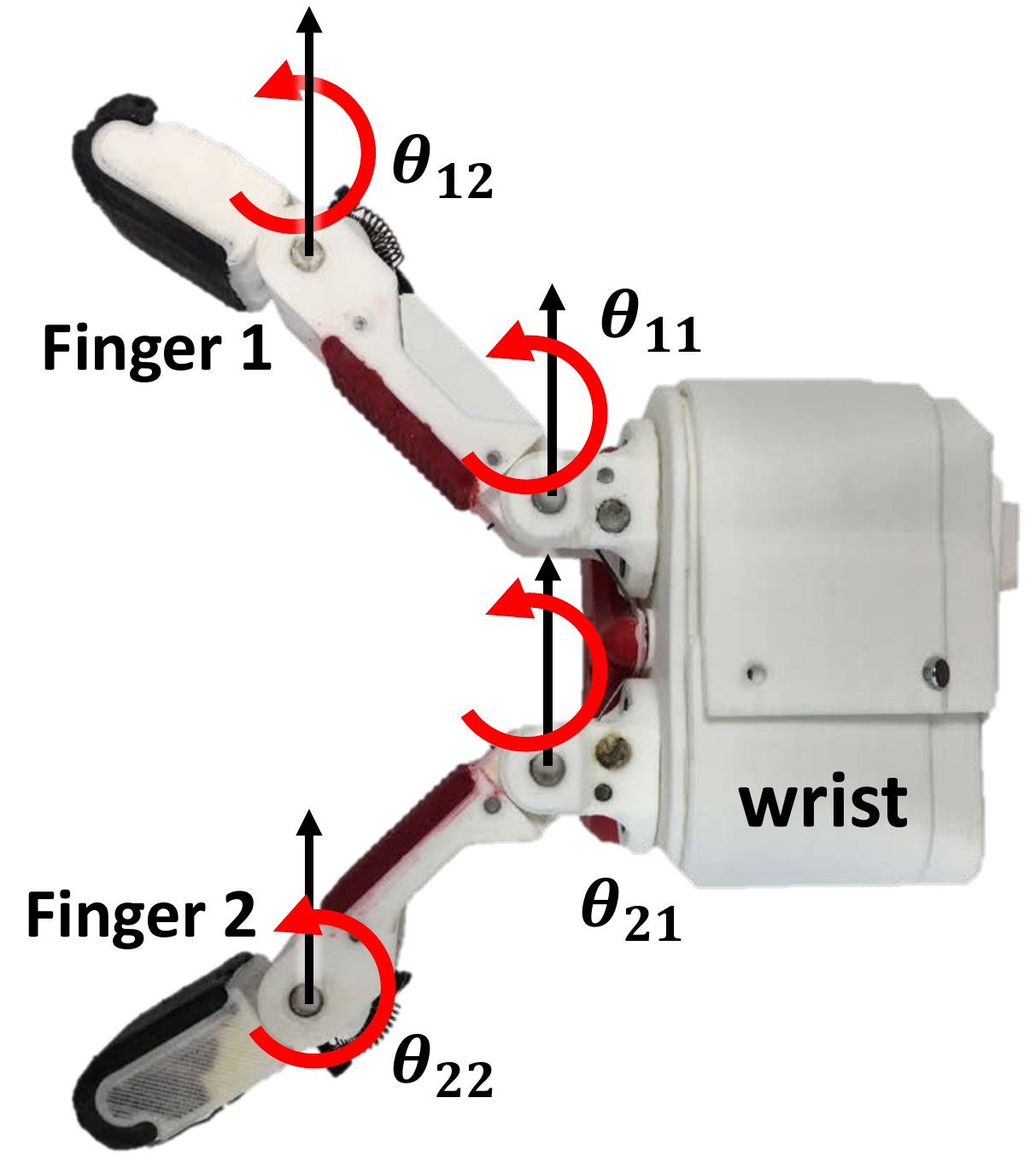}
    \vspace{-.1in}
  \caption{Adaptive hand with 2 underactuated fingers.}
  \label{fig:hand}
  \vspace{-.1in}
\end{wrapfigure}
An adaptive hand consists of a wrist and a set of fingers. The fingers are not sensorized to a level that provides reliable state information. Each finger $F$ is treated as an articulated chain and its configuration is the set of all joint angles, i.e.,  $q_F=\{\theta_{F1},\theta_{F2},...,\theta_{Fn}\}$ (see Fig. \ref{fig:hand}).  A 3D region-of-interest (ROI) is identified that contains the point cloud $P_S$ of the in-hand object and fingers. The ROI is computed based on the wrist's pose $T_C^H$ obtained from forward kinematics and the hand dimensions. {\tt ICP}, performed over the point cloud and the wrist's model, refines $T_C^H$ to compensate for errors in forward kinematics and camera calibration.

% In the case of Yale Hand T42, there are two joints and can only rotate in a horizontal plane, hence its pose is simply denoted as $\theta_F=\{\alpha_1,\alpha_2\}$, where $\alpha$ is the horizontal rotation angle

% Alternatively, in tracking scenarios rather than single image which is our case, one can also use the algorithms described in \cite{schmidt2014dart}.

%  Due to the sensor's imperfectness, noisy outlier points makes the convexity assumption of the objective function fragile and gradient descent algorithms tend to converge to local optimums. 

The next step aims to find the finger configuration, which minimizes the discrepancy between the robot hand model and the observed depth image given $P_S$. It is possible to formalize this problem as convex objective optimization and employ gradient descent algorithms to obtain the optimal pose, as in related work \cite{schmidt2014dart}. An initial estimation from the previous frame, in the context of a tracking scenario, can be good initialization for the gradient descent to converge. Nevertheless, in single image estimation, such as in this work, no such initial guess is assumed. For this reason, this paper proposes Particle Swarm Optimization ({\tt PSO}) for searching each finger configuration, inspired by prior work on human hand pose tracking \cite{qian2014realtime}. {\tt PSO} is an evolutionary process where particles interact with each other to search the parameter space. In addition to being less sensitive to local optima, it is highly parallelizable and does not require the objective function to be differentiable. This allows to formalize the cost function as minimizing the negative {\tt LCP} (Largest Common Pointset) \cite{amo_fpcs_sig_08} score computed via an efficient KDTree implementation.

Unlike human hands, the configuration space of robot hands is more constrained. It was empirically observed that instead of estimating the hand state globally in {\tt PSO}, sequentially estimating each finger's configuration leads to more stable solutions and faster convergence (with 15 particles and 3 iterations for each finger). Therefore, {\tt PSO} was applied to each finger separately to estimate its configuration starting from the finger closest to camera. Each {\tt PSO} particle is a vector representing the current finger configuration $q_F$ and the swarm is a collection of particles. Initially, particles are randomly sampled and their velocities are initialized to zero. In each generation, a particle’s velocity is updated as a randomly weighted summation of its previous velocity, the velocity towards its own best known position, and the velocity towards the best known position of the entire swarm.

The cost function evaluation is given in Alg. \ref{alg_costfunction}. The inputs are the finger configuration $q_F$, which will be evaluated, the hand region point cloud $P_S$ and finger model point cloud $P_F$. In lines 2 - 5, a penalty is assigned to cases when fingers have collisions. It returns a score that is linearly dependent on the penetration depth $d$ to encourage particles to move to a more promising parameter space that satisfies collision avoidance. The $\lambda_c$ parameter is a penalization term and is arbitrarily assigned to a very large value. $P_S$ is first transformed into the finger frame using forward kinematics and $q_F$. A KDTree is built on the transformed $P_S$ to compute the {\tt LCP} score with the finger model cloud efficiently. 

\vspace{-.15in}
\import{alg/}{costfunction}
\vspace{-.15in}

The single shot hand state estimation is implemented for parallel execution in C++. This component can also be very useful for tracking approaches \cite{schmidt2014dart, cifuentes2016probabilistic} as initialization or re-initialization.

% The entire PSO finger pose estimation process is implemented to be parallel in C++. Although in this work the exhibited hand T42 only has two fingers, we note that the same approach can be extended to more complex adaptive hands by applying PSO to all fingers at once. In real in-hand manipulation scenarios, the system can also be fused seamlessly with articulated tracking methods such as \cite{schmidt2014dart,cifuentes2016probabilistic} for tracking initialization or recovering from lost or kidnapped tracking problems. This, however, is beyond the scope of single image in-hand object pose estimation in this work.

\subsection{Object Pose Hypotheses Generation and Clustering}

Once the full hand state $x_H$ is available, $SDF$ (Signed Distance Field) is computed for the hand. All  $P_S$ points with signed distance below a threshold $SDF(p,x_H)<\epsilon$ are eliminated ($\epsilon=3 \ mm$ in the accompanying experiments). The remaining point cloud $P_O$ is now assigned to the object. The new goal is to register the object mesh $M_O$ against the point cloud $P_O$, despite the imperfections of $P_O$ due to sensor noise, occlusions or errors in the hand state estimate.

% Existing 3D registration methods such as \cite{drost2010model} are based on local geometry features followed by a voting scheme. This nature makes them sensitive to point cloud density, which becomes problematic under severe occlusions where non-uniform distribution of point cloud or poor depth sensing quality around edge areas occurs. Others such as \cite{zhou2016fast} designs some robust cost function and the alignment solution is found by gradient descent optimization. Under severe occlusions, only limited features and correspondences can be extracted on the small point cloud segment which makes the designed cost function collapse more easily. It has been found that for our target setup \textit{Super4PCS} is more appropriate.By sampling various 4-points co-planar bases from source cloud, \textit{Super4PCS} then searches for their corresponding congruent sets in the target cloud in a \textit{RANSAC} manner and generates a pool of rigid alignment hypothesis along with their scores measured by {\tt LCP}.

\begin{wrapfigure}{r}{0.2\textwidth}
    \vspace{-.15in}
  \centering
  \includegraphics[width=0.15\textwidth]{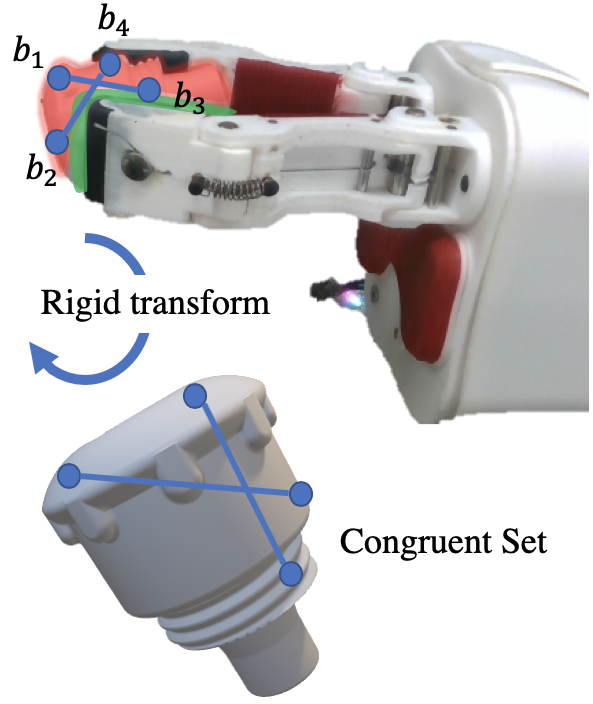}
      \vspace{-.1in}
  \caption{A 4-point base is heuristically sampled. For a congruent set on the model a candidate transform is defined.}
%      \vspace{-.2in}
  \label{fig:matching}
\end{wrapfigure}

This paper builds upon prior work for hypotheses generation \cite{mellado2014super, mitash2018robust}. It samples sets of 4-point, co-planar bases on the object's point cloud ($P_O$), and searches for congruent sets on the object model ($M_O$) to provide a pool of rigid alignments (Fig. \ref{fig:matching}). Bases can be sampled randomly \cite{mellado2014super} or given the stochastic output of a {\tt CNN} \cite{mitash2018robust}. To limit the number of samples, while maximizing the chances of sampling a valid base (where all points belong to the object), this work proposes sampling heuristics given the hand state. 

%The heuristics come from a probability model based on a \textit{Euclidean %Distance Transform}, which is easily computed from the hand's signed %distance field.

% In this work, we build on top of \textit{Super4PCS} and propose a heuristics sampling guided congruent sets sampling process inspired by \cite{mitash2018robust}, where the heuristics in our case come from a probability model based on \textit{Euclidean Distance Transform} and can be computed effortlessly thanks to our previous estimation of hand states along with the computed signed distance field.
% Given the the object model cloud $P_M$ and object related cloud segment $P_O$ after removing hand region cloud from $P_S$, the proposed method follows the principles of heuristics guided base sampling technique on $P_O$, a congruent set retrieval process on $P_M$, followed by a randomized alignment process between any two corresponding congruent sets, where a base $B:=\{b_1,b_2,b_3,b_4\}$ is defined as a set of co-planar 4-points. 

The base sampling process is given in Alg. \ref{alg_base_sample}, where inputs are the object point cloud $P_O$, heuristics $\pi$ and a hash map $PPF_M$ of Point Pair Features (PPF) \cite{drost2010model} of the model $M_O$. The hash map $PPF_M$ is precomputed. It counts the number of times a discretized PPF feature appears on $M$. The PPF for any two points on $M_O$ is given by:

\vspace{-0.25in}
\begin{align*}
    PPF\left(\mathbf{p}_{1}, \mathbf{p}_{2}\right)=\left(\|\mathbf{p_1p_2}\|_2, \angle\left(\mathbf{n}_{1}, \mathbf{d}\right), \angle\left(\mathbf{n}_{2}, \mathbf{d}\right), \angle\left(\mathbf{n}_{1}, \mathbf{n}_{2}\right)\right)
\end{align*}
\vspace{-0.25in}

\noindent where $n_1$ and $n_2$ are point normals and $d$ is the distance between the points. This avoids outliers from $P_{O}$. For sampling one base, 4 points are sampled incrementally by using a heuristic score associated with every point on the point cloud $P_O$.  The heuristic score follows an exponential distribution of the \textit{Euclidean Distance Transform} of each point, which is computed from the hand's signed distance field $SDF$:

\vspace{-0.25in}
\begin{align*}
    \pi (p_i) \propto 1-exp(-\lambda SDF(p_i;x_H)).
\end{align*}
\vspace{-0.25in}

\noindent where $\pi (p_i)$ returns a point's probability to be sampled. The probability distribution of all the points on the object cloud $P_O$ are normalized and denoted as $\pi$. Points further away from the hand are more likely to belong to the object and are prioritized. To balance exploitation and exploration, a discounting factor $\gamma=0.5$ decays the heuristic when a point is sampled. The discounting generates more dispersed and promising pose hypotheses. 

\vspace{-.15in}
\import{alg/}{base_sampling}
\vspace{-.15in}

The sampling ensures that the 4 points are co-planar given a small threshold (Line \ref{ifcoplanar}). Base sampling is repeated until a desired number of bases is achieved. Given a base $B$, its congruent set on the object model is retrieved by hyper-sphere rasterization \cite{mellado2014super}. Alignment between the matching bases can be solved in a least square manner \cite{mellado2014super}. This returns a set of object pose hypotheses along with their {\tt LCP} score. Base sampling and alignment are executed in parallel.

The large number of pose candidates generated often contains many incorrect or redundant poses. Clustering in SE(3) is performed to group together similar poses and reduce the size of the hypotheses set. Similar to prior work \cite{mitash2017improving}, a fast and effective technique is adapted for this step: a round of coarse grouping is performed in $R^3$ via \textit{Euclidean Distance Clustering}. Then, each group is split by clustering according to the minimal geodesic distance along $SO(3)$:

\vspace{-0.2in}
\begin{align*}
    d(R_1,R_2)=arccos(\frac{trace(R^T_1R_2)-1}{2}).
\end{align*}
\vspace{-0.2in}

Different from prior work \cite{mitash2017improving}, however, rather than using K-means, which can be computationally expensive, the new hypotheses are formed by the poses with the highest {\tt LCP} score per cluster and refined by {\it Point-to-Plane} {\tt ICP} \cite{chen1992object}. After {\tt ICP}, some candidates may converge to the same pose and are merged. The top $k$ hypotheses (empirically set to 100)  with the highest {\tt LCP} score are kept to improve computational efficiency.

\subsection{Pose Hypothesis Pruning and Selection}

%The pose hypothesis generation process implicitly takes into account %geometric information coming from the point cloud. However, for the task %of in-hand object pose estimation where severe occlusions occur %frequently, limited object visibility makes such geometric information %insufficient. To overcome this issue,

%This can be performed efficiently by simply looking up the signed distance value of each point in the object model cloud in the frame of the hand computed as a chained transformation $T^H_O=T^H_C T^C_O$, where $T^H_C$ is the transformation from camera to hand frame as discussed in Section \ref{sec_hand_pose}, $T^C_O$ is the object pose hypothesis.% 

% it can happen that the remaining pose hypothesis with the highest {\tt LCP} score is not the true positive due to depth sensor noise. 

Physical reasoning is leveraged to further prune false hypotheses via collision checking and scene-level occlusion reasoning. Physical consistency is imposed by checking if the object model collides beyond certain depth with the estimated hand state, or if the object is located above certain distance from the hand mesh surface, indicating that the hand is not touching the object. This process can be performed efficiently by utilizing the hand state and its $SDF$. 

Ambiguities might still arise due to several pose candidates achieving similar {\tt LCP} score with the object under high occlusions. Any (non-corrupted) observation of a non-zero depth indicates that there is nothing between the observed point and the camera, up to some noise threshold and barring sensor error \cite{ganapathi2012real}. This scene-level reasoning is adapted by comparing the accumulated pixel-wise discrepancy between the observed depth image and the rendered one (computed via {\it OpenGL} using both the estimated object pose and hand state). Based on this rendering score, the top $1/3rd$  of pose hypotheses are retained. The final optimal pose is selected from this set according to the highest {\tt LCP}.

%  Hereby, we impose this underlying nature contained exclusively in depth image
 
% Since different scenes can result in different distributions in a depth image due to sensor range or noise, in this step we do not set any hard threshold for pose hypothesis rejection as we did in physics reasoning part, but rather select the pose hypothesis which has minimal difference $D$.

%specify the number of base we sample (3)%

\begin{figure*}[t!]
\centering
\includegraphics[width=0.95\textwidth]{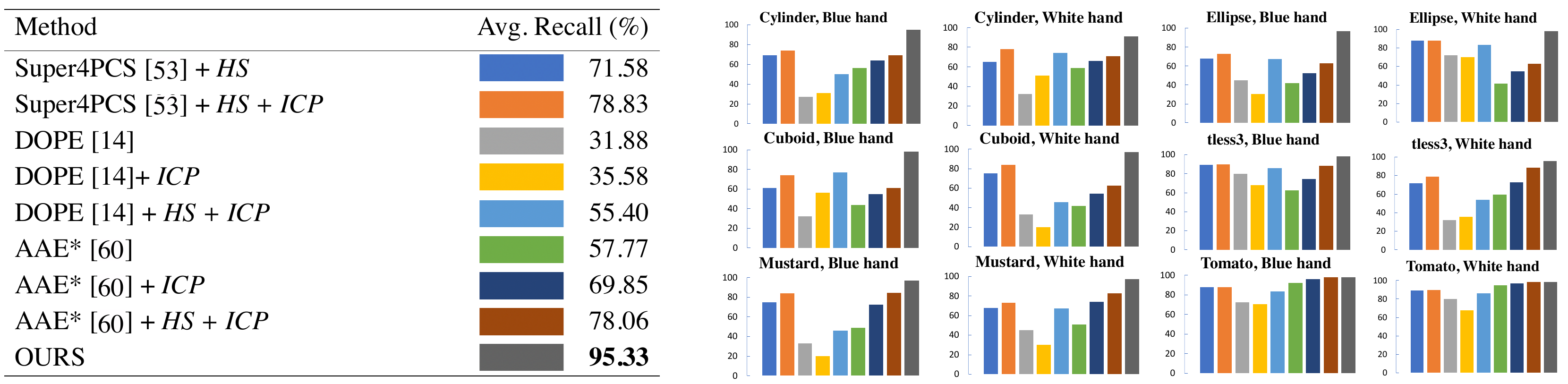}
\vspace{-0.1in}
\caption{Comparison on simulation dataset. For the table, \textit{+HS} implies using the proposed {\tt PSO} hand pose estimation to remove the hand related cloud from the scene, \textit{+ICP} implies applying Point-to-Plane ICP for pose refinement.}
\label{simulation_results}
\vspace{-0.25in}
\end{figure*}

%% file: alg/costfunction.tex
\begin{algorithm}[h]
\label{alg_costfunction}
\caption{{\sc COST\_FUNCTION\ ($q_F$, $P_S$, $P_F$)} }
$P_S^{finger} \gets $ transform $P_S$ to finger frame using forward kinematics and $q_F$\; 

\For{any other finger $Q_F$}
{   
    \tcc{collision penetration depth (negative)}
    $d \gets collisionCheck(P_F,Q_F)$\;
    \If{$d < \epsilon$}
      {
    	return $-\lambda_c -\lambda_cd$\;
      }
}

$kdtree(P_S) \gets$ build kdtree from $P_S^{finger}$\;
$LCP \gets 0$\;
\For{$each \ p_F \in P_F$}
{
    $p_{nei} \gets kdtree(P_S).findNearestNeighbor(p_F) $\;
    \If{$||p_{nei}-p_F||<\epsilon$ and $normal(p_{nei}) \cdot normal(p_F) > \delta $}
      {
    	$LCP \gets LCP + 1$\;
      }
}

return $-LCP$\;
\end{algorithm}

%% file: alg/base_sampling.tex
\begin{algorithm}[h]
\label{alg_base_sample}
\caption{{\sc SAMPLE\_ONE\_BASE\ ($P_O$, $\pi$, $PPF_M$)} }
% $B = \{\}$ \;
$b_1$ $\gets$ sample a point from $P_O$ according to $\pi$ \;
$B \gets \{b_1\} $ \;
\For{$p \in P_O$}
{
$f_1 \gets PPF(p, b_1)$ \;
\If {$PPF_M [ f_1 ] ==  \O $}{
$\pi(p) \gets 0$ \;
}
}
\For{$i\gets0$ \KwTo  max\_iter }
{
  $b_2$, $b_3$ $\gets$ sample two different points from $P_O$ according to the updated distribution $\pi$ \;
  $\pi(b_2)$ $\gets$ $\gamma$ $\pi(b_2)$ \;
$\pi(b_3)$ $\gets$ $\gamma$ $\pi(b_3)$ \;

  $f_{23} \gets PPF(b_2,b_3)$ \;
  \If {$PPF_M [ f_{23} ] \neq  \O $ and $\angle  (\overrightarrow{b_1b_2},\overrightarrow{b_1b_3})>\delta$}{
    $B \gets B \cup \{b_2, b_3\}$ \;
    break \;
  }
}

\For{$i\gets0$ \KwTo  max\_iter }
{
  $b_4 \gets $ sample a point from $P_O$ according to the updated distribution $\pi$ \;
  $\pi(b_4)$ $\gets$ $\gamma$ $\pi(b_4)$ \;
  \If{$distance(plane(b_1,b_2,b_3),b_4)>\epsilon $ \label{ifcoplanar}} 
  {
	continue \;  
  }
  $f_{24} \gets PPF(b_2, b_4)$, $f_{34} \gets PPF(b_3, b_4)$ \;
  \If{ $PPF_M [ f_{24} ]\neq \O$ and 
        $PPF_M [ f_{34} ]\neq \O$}
  {
    $B \gets B \cup b_4$ \;
	break \;
  }
}
return $B$\;
\end{algorithm}

%% file: sections/experiments.tex
% \import{tables/}{sim_results}

This section evaluates the proposed approach and compares against state-of-the-art single-image pose estimation methods on in-hand objects. Note the difference with tracking methods \cite{schmidt2015depth,schmidt2014dart,cifuentes2016probabilistic,issac2016depth}, since here the 6D object pose is recovered from a single static image without dependency on previous frames. To the best of the authors' knowledge, there are no relevant datasets in the literature beyond those for objects in human hands \cite{issac2016depth}. A benchmark dataset is developed that includes both simulated and real world data for in-hand object pose estimation with adaptive hands and will be released publicly.

\subsection{Experimental Setup}
\begin{figure}[thpb]
  \centering
  \vspace{-0.15in}
  \includegraphics[width=0.95\columnwidth]{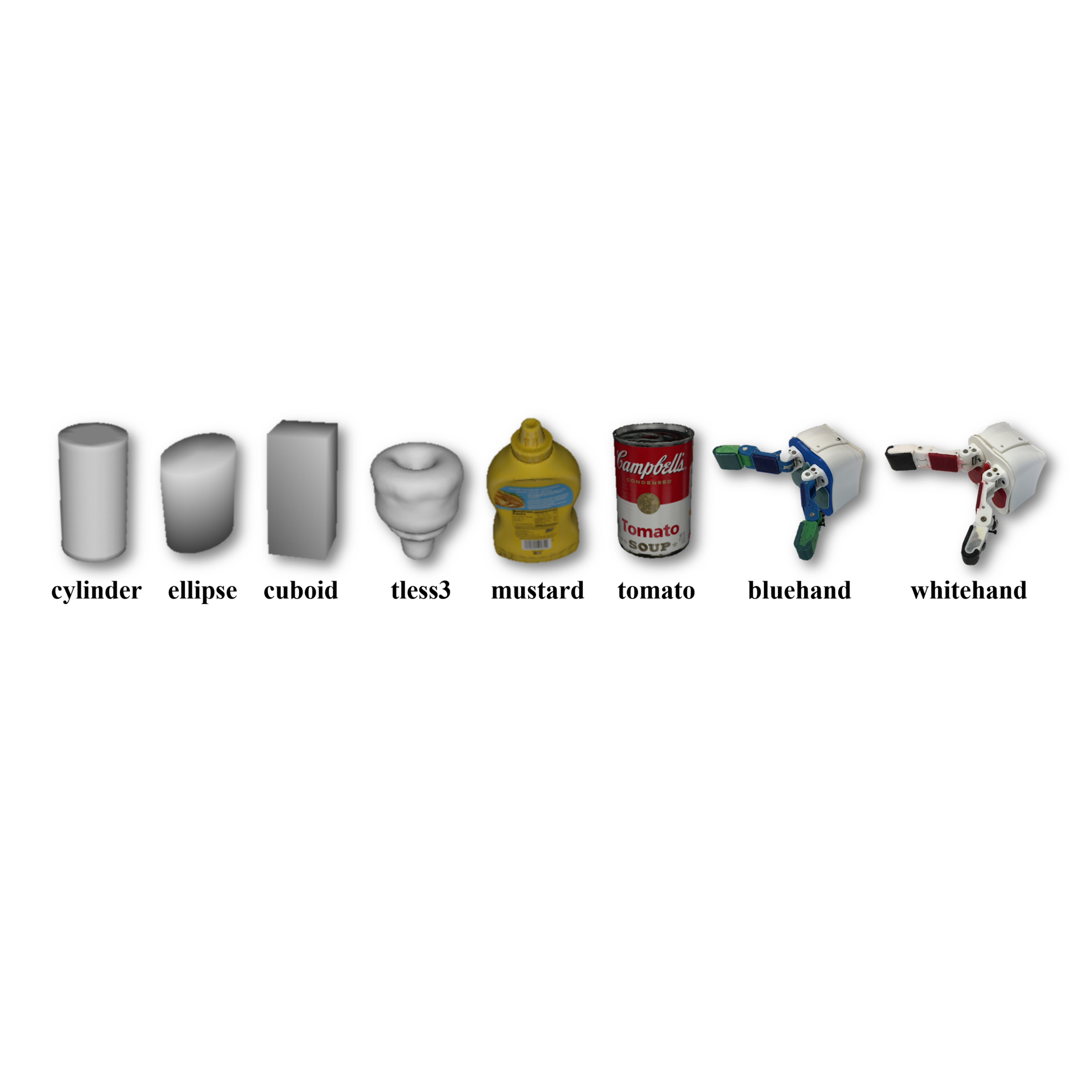}
  \vspace{-0.15in}
  \caption{Mesh of objects used: A cylinder with diameter 0.035 $m$ and length 0.064 $m$, an ellipsoid with length 0.064 $m$, a cuboid with side length 0.03 $m$ and length 0.064 $m$, an industrial object \#3 from T-LESS dataset \cite{hodan2017tless}, a mustard bottle and tomato soup can from YCB dataset \cite{calli2015ycb}. Right 2 images: Yale T42 adaptive hands painted in blue and white.}.
  \vspace{-0.15in}
  \label{objects}
\end{figure}

The setup consists of a robot manipulator (Yaskawa Motoman) and a Yale T42 adaptive hand (Fig.~\ref{fig:hand}), which was 3D printed based on open-source designs. Objects considered for in-hand manipulation were picked to evaluate the robustness of estimation. As shown in Fig.~\ref{objects}, the selected set is a mix of objects: with and without texture or geometric features.

% The performance of our approach is quantitatively and qualitatively evaluated on both simulation and real data and outperforms other state-of-art methods by a considerable margin, indicating that general 6D object pose estimation framework can hardly meet the demand in real robot in-hand manipulation scenarios. 

All experiments are conducted on a standard desktop with Intel Xeon(R) E5-1660 v3@3.00GHz processor. For the comparison to deep learning methods, neural network inference is performed on a NVIDIA Tesla K40c GPU.

\subsection{Evaluation Metric}

The recall for pose estimation is measured based on the error given by the
{\tt ADI} metric \cite{hinterstoisser2012model}, which measures the average of point distances between poses $T_1$ and $T_2$ given an object mesh model $M$:

\vspace{-0.2in}
\begin{align*}
e_{ADI} (T_1,T_2) = \text{avg}_{p_1 \in M} \text{min}_ {p_2 \in M} || T_1(p_1,M) - T_2(p_2,M)||_2,
\end{align*}
\vspace{-0.2in}

\noindent where $T(p,M)$ corresponds to point $p$ after applying
transformation $T$ on $M$. Given a ground-truth pose $T^g$, a true
positive is a returned pose $T$ that has $e_{ADI}(T, T^g) < \epsilon $, where $\epsilon$ is a tolerance threshold. $\epsilon$ is set to 5 $mm$ in all experiments except in recall curves, to evaluate the applicability of different methods for precise in-hand manipulation scenarios.

\subsection{Simulation Dataset and Results}

% since this work aims at any-time single image 6D object pose estimation during %the entire within-hand manipulation process, in which a stable grasp is not %always a true assumption

Simulated RGB-D data were generated by placing a virtual camera at random poses around the model of the hand. Poses are  sampled from 648 view points on spheres of radius 0.3 to 0.9 $m$ centered at the hand. To generate each data point, an object is placed at a random pose between the fingers. The two articulated-fingers are closed randomly until they touch the object, verified by a collision checker. Physical parameters, such as friction, gravity or any grasping stability metric are deliberately not employed since this work aims at any-time single image 6D object pose estimation during the entire within-hand manipulation process, in which a stable grasp is not always a true assumption. By randomizing the object pose relative to the hand, the dataset is able to cover various in-hand object poses that can occur during an in-hand manipulation process. For the adaptive hand, two colors are chosen. The blue hand differs from any object color used in the experiments whereas the white hand resembles texture-less objects and evaluates robustness to lack of texture. In addition to the RGB-D data, ground-truth object pose and semantic segmentation images are also obtained from the simulator. For each combination of the 6 objects and the 2 adaptive hands, 1000 data points are generated, resulting in 12000 test cases.

% By showing/hiding the adaptive hand during rendering, we are also able to compute the occlusion ratio of the object by comparing the segmentation images with/without adaptive hand.

% Super4PCS \cite{mellado2014super} shares the most similarity with ours in finding congruent sets and compute their relative rigid transformation. 

Fig.~\ref{simulation_results} reports the recall for pose estimation on the synthetic dataset.  When {\tt Super4PCS} is directly applied to the entire point cloud, outlier points that do not belong to the object are often sampled, leading to poor results (5.83 \%). On introducing the proposed {\tt PSO} hand state estimation (HS) and thereby eliminating the hand points from the scene, points belonging to the object are more likely to be sampled, which dramatically improves the performance of (Super4PCS+{\it HS}). Recent state-of-the-art learning-based approaches are also evaluated. DOPE \cite{tremblay2018deep} trains a neural network to predict 3D bounding-box vertices projected on the image and recovers 6D pose from them via Perspective-N-Point (PnP), which has shown to outperform PoseCNN \cite{xiang2017posecnn} on the YCB dataset. To eliminate the domain gap from the scope of evaluation, the training and test data were generated in the same simulator and domain randomization was utilized as suggested \cite{tremblay2018deep}. AAE \cite{sundermeyer2018implicit} is another learning-based method that trains an autoencoder network to embed object 3D orientation information using extensive data augmentation and domain randomization techniques. It has been shown to be successfull on textureless objects and achieved state-of-art results on the T-LESS dataset \cite{hodan2017tless}. This approach is only able to predict 3D orientation. The translation is based on the output of another object detection network. For the scope of this evaluation {\it ground-truth} bounding-box were provided as input to AAE \cite{sundermeyer2018implicit}.

% Due to the ambiguous appearance of cylinder and cuboid on image, as well as occlusions introduced by robot hand, it is difficult for learning-based approaches to predict precise 6D pose of in-hand object, which aligns with the findings in \cite{hodan2018bop}. 

 A dramatic performance improvement is observed for all methods, when the proposed {\tt PSO} hand state estimation is utilized to remove the hand related point cloud from the scene. This proves the significance of additionally estimating the robot hand state for in-hand 6D object pose estimation.

\import{tables/}{real_results}

\begin{figure}[thpb]
\vspace{-0.1in}
  \centering
  \includegraphics[width=0.95\columnwidth]{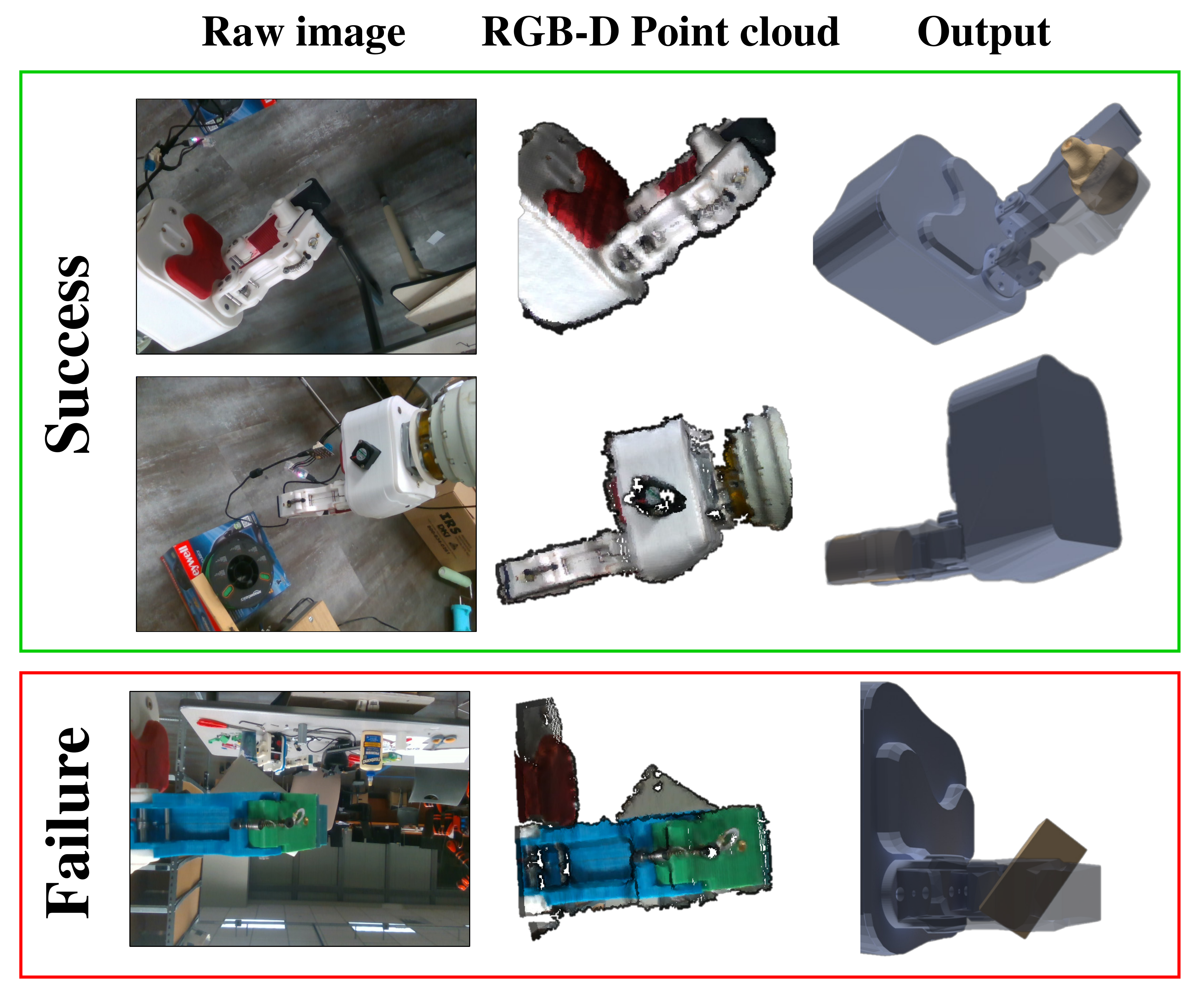}
  \vspace{-0.2in}
  \caption{Qualitative results for the proposed approach showing success and failure cases under challenges like occlusion and symmetry.}.
  \label{qual}
\vspace{-0.34in}
\end{figure}

\subsection{Real Dataset and Results}

The real dataset contains 986 snapshots of 2 Yale T42 hands holding 4 types of objects including cylinder (295), ellipse (239), cuboid (187) and tless3 (265). All the objects and the adaptive hands are 3D printed. Similar to the setting in simulation, the adaptive hands are painted in two colors: blue-green and white. The images are collected with an Intel RealSense SR300 RGB-D camera and the ground-truth poses are manually annotated using a GUI developed by the authors. Before each image is taken, objects are grasped randomly and the adaptive hand performs a random within-hand manipulation. Due to the small size of objects relative to the hand, severe occlusions occur frequently, as exhibited in Fig. \ref{qual}. 

\begin{figure}[h]
\vspace{-0.1in}
    \centering
    \includegraphics[width=0.23\textwidth]{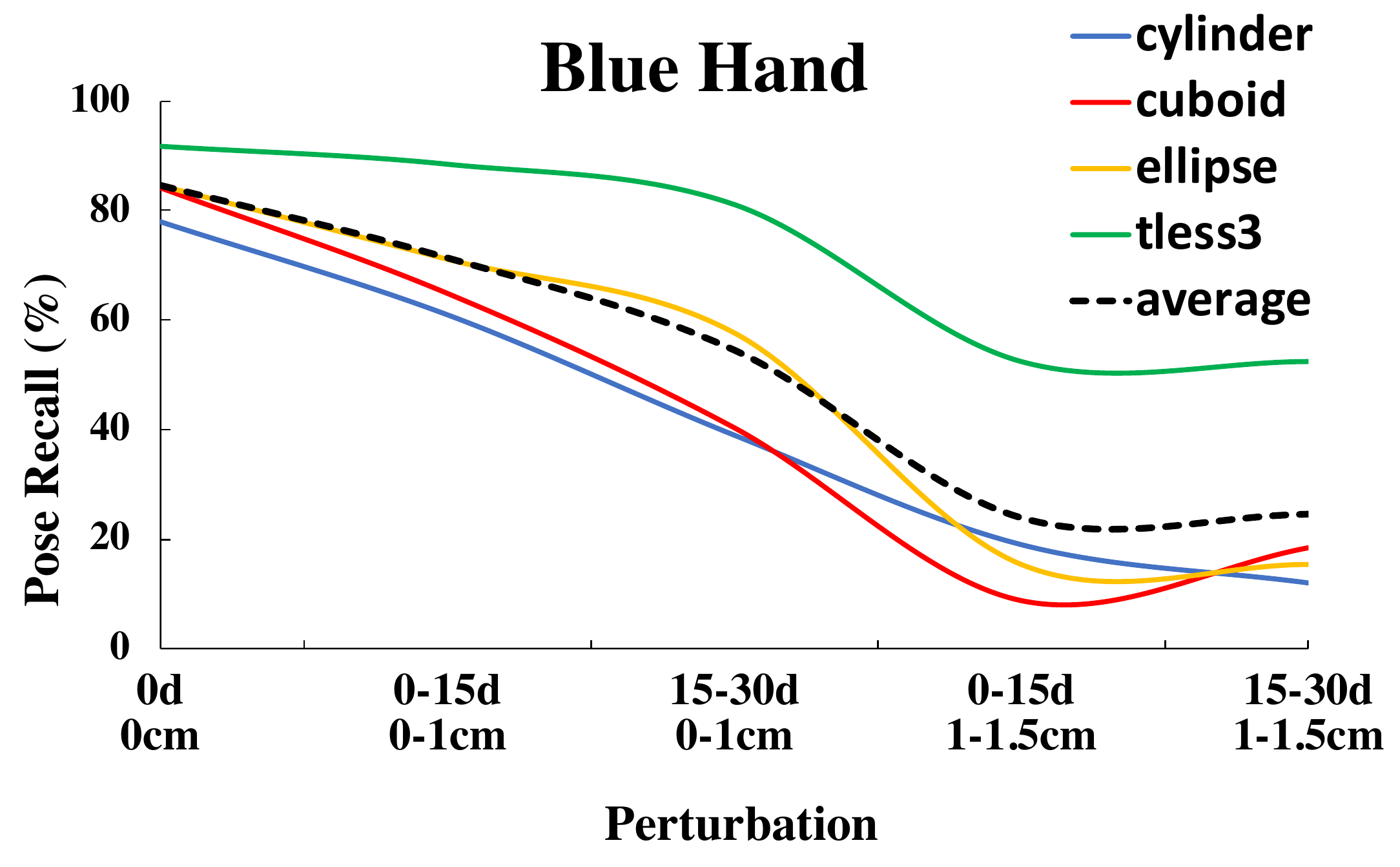}
    \includegraphics[width=0.23\textwidth]{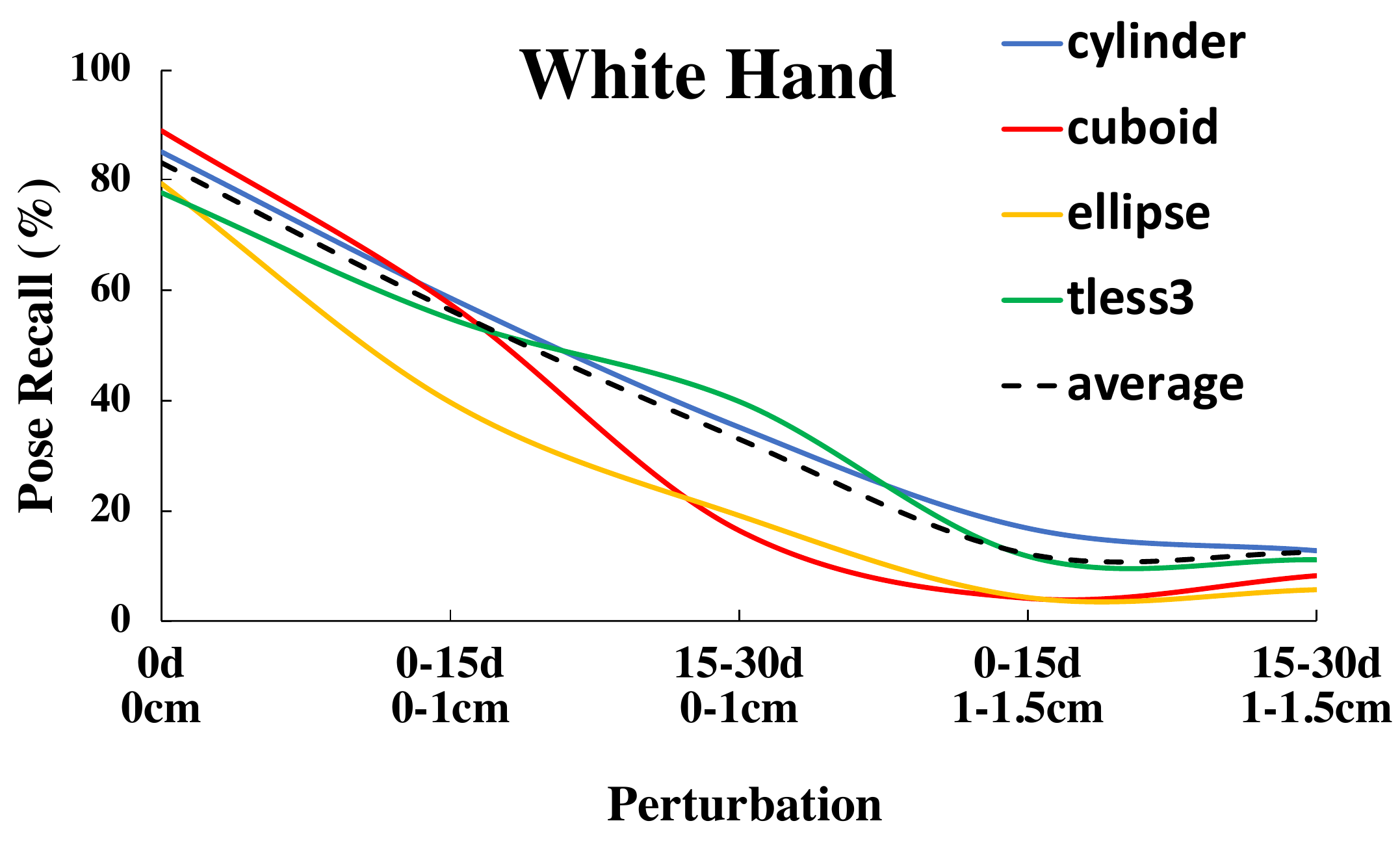}
    \vspace{-0.1in}
    \caption{Pose recall of \cite{choi2016using} on the real dataset. As the approach requires initialization, it is evaluated over perturbations on the ground-truth pose.}
    \label{fig:choi}
\vspace{-0.1in}
\end{figure}

% \begin{figure}[t]
%     \centering
%     \includegraphics[width=0.5\textwidth]{fig/realdata_recall_curve.pdf}
%     \vspace{-0.3in}
%     \caption{Recall-threshold curves for evaluated techniques.}
%     \label{fig:real_recall}
% \vspace{-0.2in}
% \end{figure}
 
Table \ref{real_results} presents results on real data. Given the large appearance gap between synthetic training data and real scenarios, and the presence of textureless objects, the performance of DOPE does not translate well, and thereby was dropped from the table. AAE \cite{sundermeyer2018implicit} was robust to some of these challenges and given the ground-truth bounding-boxes, it could predict the correct rotation in some cases. An additional related work \cite{choi2016using} was evaluated on real data. It was developed to perform pose estimation for in-hand objects during robot manipulation. It assumes the initial object pose does not change much upon grasping and serves as an initialization for ICP. To evaluate this approach, pose initialization is provided by perturbing the ground-truth pose. Fig.~\ref{fig:choi} shows how the performance of this approach varies with the perturbation. Our proposed approach outperforms the best-case (small perturbation) of \cite{choi2016using} even though pose initialization is not provided to our system.

\import{tables/}{ablations}

\subsection{System Analysis}

Fig.~\ref{qual} exhibits examples of the output from the proposed approach on real data where severe occlusions occur and additional challenge is introduced by virtue of the noise in consumer-level depth sensor. Table.~\ref{ablation} shows the ablation study where the recall percentage for the object pose ($e_{ADI}<5mm$) is measured by incrementally adding the critical proposed components. \import{tables/}{runtime} Table~\ref{runtime} presents the overall and the decomposition of the running time for each component of the proposed pipeline, when tested on real data. $Misc$ includes transformations, building KDTree, etc. Given the parallel implementation, the proposed technique requires a relatively short amount of time to perform a single image pose estimation without any initialization such as in tracking.

%% file: tables/real_results.tex
\begin{table*}[!ht]
\begin{minipage}{0.65\textwidth}
\begin{small}
\begin{tabular}{p{3.8cm}p{1cm}p{1cm}p{1cm}p{1cm}p{1cm}p{1cm}}
\hline 
Method & Modality & cylinder & cuboid & ellipse & tless3 & Avg. \\\hline
Super4PCS \cite{mellado2014super} + {\it HS} & Depth & 52.49 &	43.85 &	62.64 &	62.64 &	55.41 \\
Super4PCS \cite{mellado2014super} + {\it HS+ICP} & Depth & 70.51 &	43.85 &	54.81 &	78.49 &	61.92 \\
$AAE^*$ \cite{sundermeyer2018implicit} & RGB & 11.19 &	8.56 &	15.92 &	40.38 &	19.01 \\
$AAE^*$ \cite{sundermeyer2018implicit} + {\it ICP} & RGBD & 43.39 &	22.99 &	27.35 &	55.85 &	37.40 \\ 
$AAE^*$ \cite{sundermeyer2018implicit} + {\it HS+ICP} & RGBD & 41.02 &	29.41 &	29.80 &	81.89 &	45.53 \\ 
OURS & Depth & \textbf{87.12} & \textbf{72.19} & \textbf{80.82} & \textbf{93.96} & \textbf{83.52} \\ \hline
\end{tabular}
\end{small}
\end{minipage}
\hfill
\begin{minipage}{0.28\textwidth}
\includegraphics[scale=0.26]{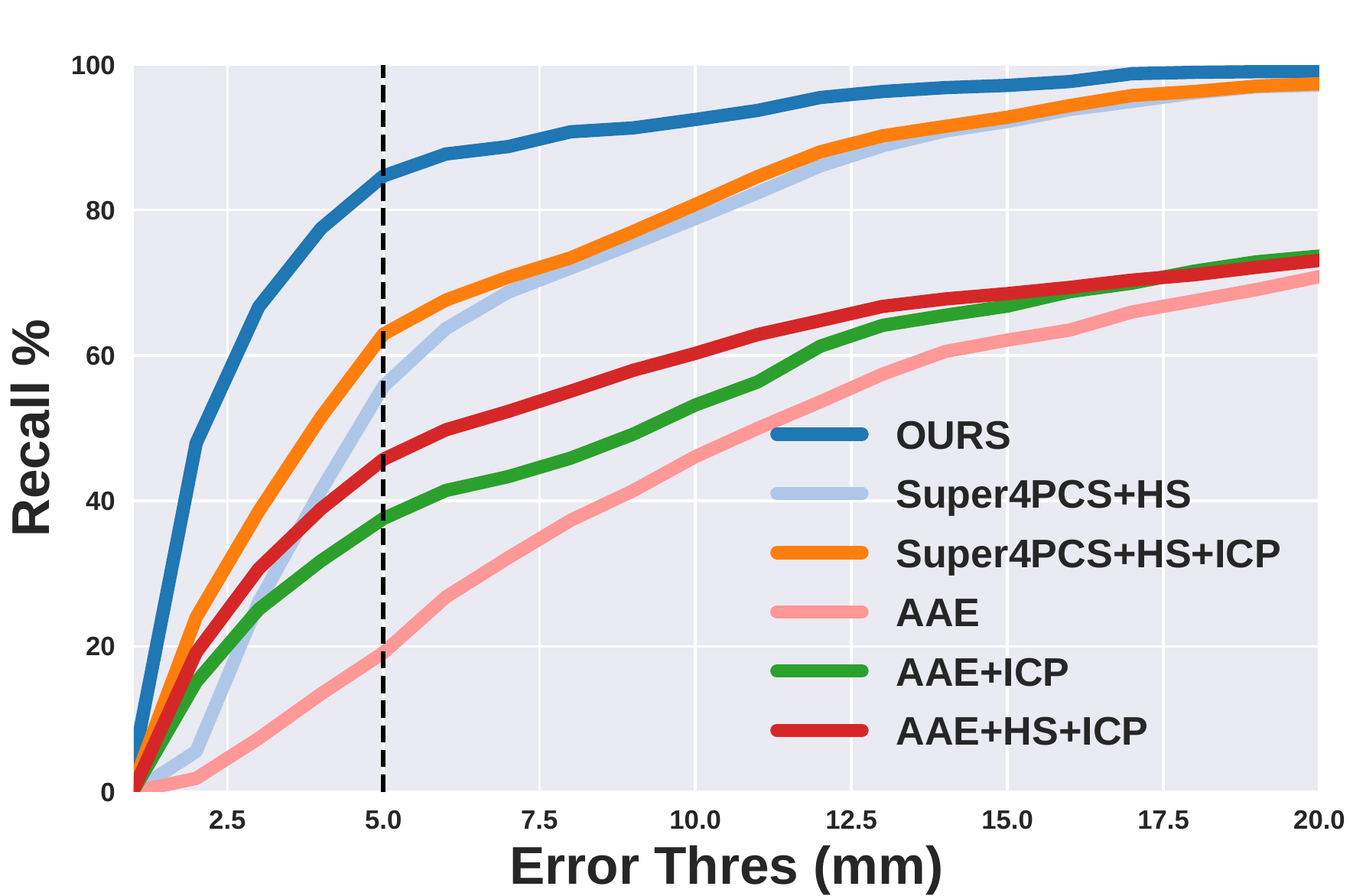}
\end{minipage}
\vspace{-.1in}
\caption{\footnotesize \textbf{Left:} Recall percentage ($e_{ADI}<5 mm$) on real data: \textit{+HS} means using the proposed PSO hand state estimation to remove the hand's point cloud, \textit{+ICP} means applying Point-to-Plane ICP at the end for pose refinement. AAE* \cite{sundermeyer2018implicit} is provided a ground-truth bounding-box.  {\bf Right:} recall-threshold curves of compared methods on real data. \vspace{-.15in}} 

\label{real_results}
\end{table*}

%Results of Choi's \cite{choi2016using} are shown in Fig. \ref{fig:choi}.

%% file: tables/ablations.tex
\begin{table}[ht]
\vspace{-.1in}
\centering
\begin{tiny}
\resizebox{0.45\textwidth}{!}
{

\begin{tabular}{lr}
\hline
Method & Mean recall (\%)\\ 
\hline\hline
Baseline & 8.44 \\
(+) {\tt PSO} handpose & 61.92 \\
(+) {\tt PPF}-constrained sampling & 75.97 \\
(+) Heuristic sampling & 79.80 \\
(+) Hypothesis pruning & 83.52 \\
\hline
\end{tabular}

}
\end{tiny}
\vspace{-.05in}
\caption{\footnotesize Ablation study of critical components in our system. Results are averaged across the entire real dataset. {\tt Baseline} refers to random base sampling on the entire scene cloud.\vspace{-.15in}}
\label{ablation}
\end{table}

%% file: tables/runtime.tex
% \begin{table}[ht]
% \vspace{-0.1in}
% \begin{wraptable}{l}{0.31\textwidth}
% % \begin{small}
% \resizebox{0.25\textwidth}{!}
% {
% \begin{tabular}{|c|r|r|}
% \hline
% \multirow{2}{*}{Component} & \multicolumn{2}{c|}{Speed (ms)} \\ \cline{2-3} 
%  & \multicolumn{1}{l|}{Mean} & \multicolumn{1}{l|}{Std} \\ \hline
% Pointcloud processing & 66.47 & 10.63 \\ \hline
% Handbase ICP & 9.15 & 3.46 \\ \hline
% Hand pose estimation & 45.98 & 3.58 \\ \hline
% Pose hypothesis generation & 90.06 & 12.00 \\ \hline
% Pose clustering and ICP & 61.41 & 11.84 \\ \hline
% Pose Hypothesis pruning & 231.47 & 62.95 \\ \hline
% Pose selection & 73.95 & 14.05 \\ \hline
% Misc & 38.17 & 10.60 \\ \hline\hline
% Total & \multicolumn{1}{l|}{616.64} & \multicolumn{1}{l|}{64.50} \\ \hline
% \end{tabular}
% % \end{small}
% }
% \caption{\small Run-time decomposition of the system on real data.}
% \label{runtime}
% \end{wraptable}
% \end{table}

% \begin{table}[h]
% \vspace{-0.1in}
\setlength{\intextsep}{2pt}%
\setlength{\columnsep}{-5pt}%
\begin{wraptable}{l}{0.25\textwidth}
% \vspace{-.15in}
% \begin{small}
\resizebox{0.25\textwidth}{!}
{
\begin{tabular}{|c|r|r|}
\hline
\multirow{2}{*}{Component} & \multicolumn{2}{c|}{Speed (ms)} \\ \cline{2-3} 
 & \multicolumn{1}{l|}{Mean} & \multicolumn{1}{l|}{Std} \\ \hline
Pointcloud processing & 66.47 & 10.63 \\ \hline
Hand wrist ICP & 9.15 & 3.46 \\ \hline
Hand pose estimation & 45.98 & 3.58 \\ \hline
Pose hypothesis generation & 90.06 & 12.00 \\ \hline
Pose clustering and ICP & 61.41 & 11.84 \\ \hline
Pose Hypothesis pruning & 231.47 & 62.95 \\ \hline
Pose selection & 73.95 & 14.05 \\ \hline
Misc & 38.17 & 10.60 \\ \hline\hline
Total & \multicolumn{1}{l|}{616.64} & \multicolumn{1}{l|}{64.50} \\ \hline
\end{tabular}
% \end{small}
}

\caption{\footnotesize Run-time decomposition of the system on real data.}

\label{runtime}
\end{wraptable}
\setlength{\intextsep}{2pt}%
\setlength{\columnsep}{10pt}%
% \end{table}

%% file: sections/conclusion.tex
% Many manipulation tasks require accurate estimation of the 6D pose of
% an object relative to the robot's end-effector upon grasping or while
% the object is being manipulated. 
% The main contributions of this project include:
% \begin{itemize}
% \item Generating
% physically realistic data in simulation; 
% \item Estimating robot hand and finger configuration through parallel search; 
% \item Performing
% registration between the segmented point cloud against the object model; 
% \item Improving the pose estimate at the scene level by
% considering physical constraints based on previously estimated relationship between hand and object, such as hand-object penetration or hand object touching interactions.
% \end{itemize}

% Our extensive experiments demonstrated both qualitatively and quantitatively the robustness and high success rate of this framework. However, current framework only uses depth information from a RGBD camera. Future work will explore the possible improvements by leveraging data-driven methods using RGB information. 

This work presents a framework for fast and robust 6D pose estimation of in-hand objects. Due to the lack of relevant datasets, both real and synthetic data will be released as a benchmark for 6D object pose estimation applied to robot in-hand manipulation. Extensive experiments demonstrate advantages of the proposed method: robustness under severe occlusions and adaptation to different objects while able to run fast as a single-image pose estimation method. Although not real time, it could be integrated with tracking-based methods to provide initialization or recovery from lost tracking.